\documentclass[a4paper, 10pt, conference]{ieeeconf}      

\IEEEoverridecommandlockouts                              

\usepackage{graphicx} 
\usepackage{amsmath} 
\usepackage{amssymb}  
\usepackage{color, colortbl}
\definecolor{Gray}{gray}{0.9}
\usepackage{multirow}
\usepackage{cite}

\title{\LARGE \bf
Passive Vibration Control of a 3-D Printer Gantry
}

\author{Maharshi A. Sharma$^{1}$ and Albert E. Patterson$^{1,2,\dagger}$
\thanks{*This work was not supported by any organization}
\thanks{$^{1}$J. Mike Walker '66 Department of Mechanical Engineering, Texas A\&M University, 3123 TAMU, College Station, Texas 77843, USA}%
\thanks{$^{2}$Department of Engineering Technology and Industrial Distribution,
        Texas A\&M University, 3367 TAMU, College Station, Texas 77843, USA }%
\thanks{$^\dagger$Corresponding Author: \tt\small aepatterson5@tamu.edu}
}

\begin{document}

\maketitle
\thispagestyle{empty}
\pagestyle{empty}

\begin{abstract}

Improved additive manufacturing capabilities are vital for the future development and improvement of ubiquitous robotic systems. These machines can be integrated into existing robotic systems to allow manufacturing and repair of components, as well as fabrication of custom parts for the robots themselves. The fused filament fabrication (FFF) process is one of the most common and well-developed AM processes but suffers from the effects of vibration-induced position error, particularly as the printing speed is raised. This project adapted and expanded a dynamic model of an FFF gantry system to include a passive spring-mass-damper system controller attached to the extruder carriage and tuned using optimal parameters. A case study was conducted to demonstrate the effects and generate recommendations for implementation. This work is also valuable for other mechatronic systems which operate using an open-loop control system and which suffer from vibration, including numerous robotic systems, pick-and-place machines, positioners, and similar.  

\end{abstract}

\section{INTRODUCTION}

In the context of robotics, where rapid iteration and adaptation are extremely important, additive manufacturing (AM) and fused filament fabrication (FFF) in particular provide a useful and straightforward way to create complex and customized components, regardless of geometric complexity~\cite{Jing2022, Kanthimathi2024}. While particularly well-adapted for complex components, simple ones can be easily and quickly made on demand, avoiding the need for custom tools or a large manufacturing space. This capability allows for rapid design modifications that facilitate the testing and optimization of robotic systems in real time, as well as the integration of direction manufacturing and repaid capabilities into other robotic systems. This, along with the relatively low machine complexity and cost of entry, reduces the barrier to entry for research and development in robotics, making advanced technologies more accessible. The ability to print highly specialized parts on demand also reduces the dependence on existing supply chains, which is particularly beneficial in remote or resource-limited environments, where traditional manufacturing might be unfeasible. Manufacturing for and with robotic systems in difficult expeditionary environments (war theaters, disaster relief, remote research stations, space exploration) will particularly benefit from this~\cite{Hu2024, Fiske2018}.  

\begin{figure*}[h!]
\centering
\includegraphics[width=1\textwidth]{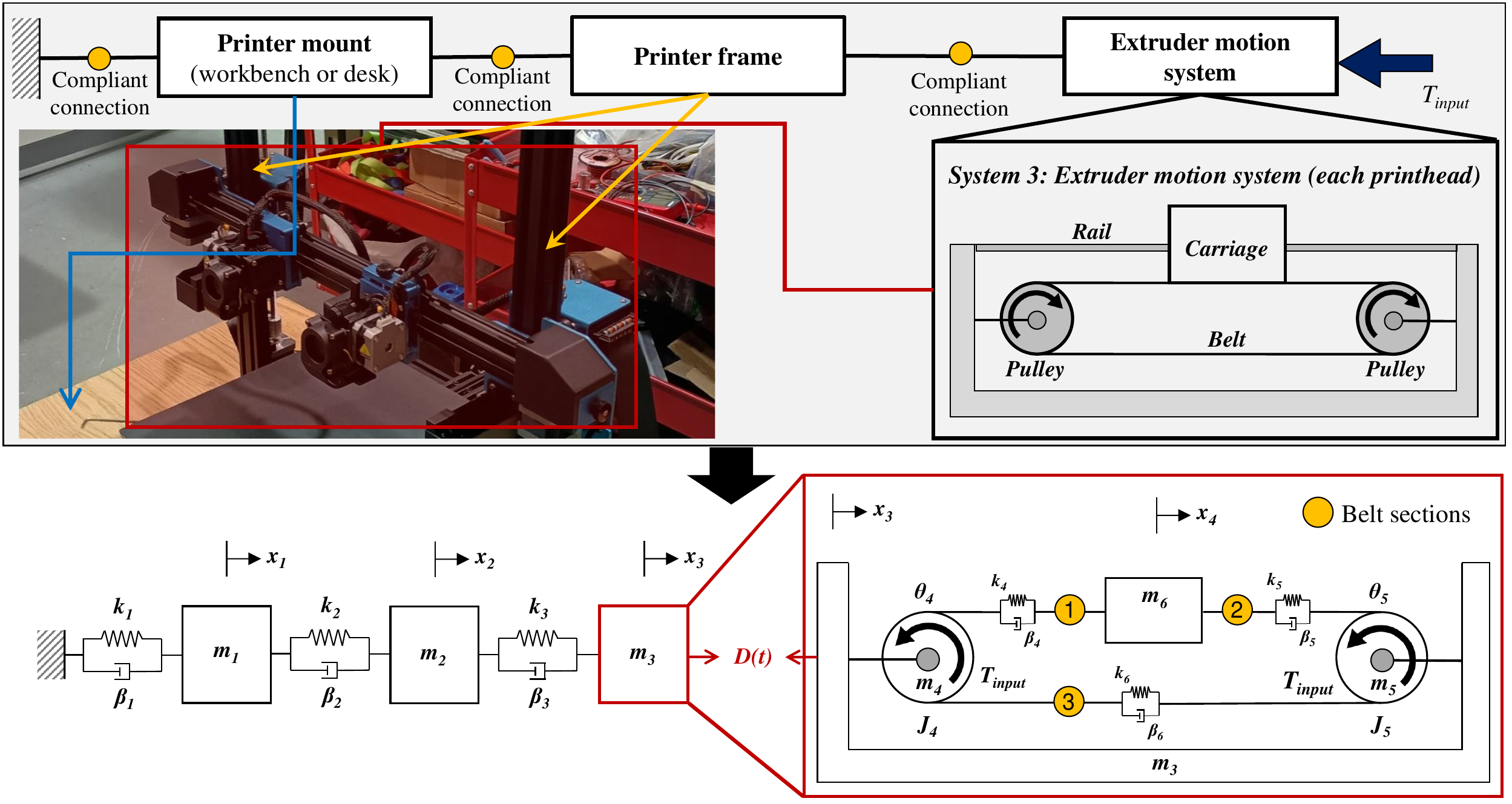}
\caption{\textit{FFF gantry system model, considering only the motion of the extrusion head and system along the X-axis. In this model, the Y-axis build plate is fixed, and the Z-axis is assummed to be a fixed distance. This is realistic for standard FFF printer configurations as shown.}}
\label{fig:system}
\end{figure*}

FFF systems can be configured in various ways, with the key requirement being the ability to position the extruder head near a secure build surface. Since the printhead is generally driven by a motor that is also contained in the system and is not directly connected to the ground, this often becomes an inverse dynamics problem with complex vibration considerations, which become worse as the deposition speed is increased. Therefore, the speed limitations on the FFF process come from vibration in the system and not from material rheology and melt properties~\cite{Go2017, Akhoundi2022, Sharma2023B, Sharma2023A}. The most common configuration is the Cartesian-based 3-axis system, although delta robots and other specialized setups are also employed, typically for research purposes. The Cartesian XYZ system remains the preferred choice for most applications due to its cost-effectiveness, ease of operation, and simplicity of maintenance. Despite extensive research on delta robots, there has been relatively little focus on the mechanics of Cartesian-based FFF systems. This work, therefore, centers on Cartesian-frame FFF machines. Mechanical compliance and vibration within these systems can introduce defects in prints, compromising part accuracy and stability, particularly at high speeds and accelerations. Although increasing print speeds is essential for reducing part costs and increasing efficiency, it also increases the risk of errors~\cite{Alexopoulou2024, LuisPrez2021, Elkaseer2020}. 

One of the most significant vibration-induced errors in FFF machines is the mismatch of the extruder nozzle with the calculate position on the build surface. The primary source of this overshoot is mechanical compliance and vibration in the print frame and drive belts. Control systems, both open-loop and closed-loop, offer potential solutions. However, closed-loop control systems may limit printing speed and their dynamic models are often time-varying, making them difficult to measure and predict. Closed-loop systems for FFF machines are also often difficult to implement, as they require the use of direct encoders, which limit the print speed. The use of vision systems would help to overcome this, but current vision systems are not accurate or fast enough for the scale of printing needed. Replacement of belts with drive screws could help with some of these vibration issues but would also create other dynamic problems and severely limit the speed of printing. 

Given that the standard FFF systems are open-loop and are controlled using a dead-reckoning method that relies on radial encoders in the driver motors, a passive open-loop control system could be a good solution. This is particularly true since the system dynamics makes it very difficult to integrate a direct control method into the system. Some tuned passive control systems could also be made semi-active or hybrid by integrating active damping or stiffness into the passive system added to the printer. The current study explores the use of a passive spring-mass-damper system that serves as a counter to vibration in the system. This work is useful for system design and contributes to robotic technology by exploring the effects of such a control system and by extracting information about proper spring-mass-damper system parameters. This work focuses on the gantry system, which is the main driving portion of the FFF system, leaving the effects of printbed motion for future work. 

\renewcommand*{\arraystretch}{1.15}
\begin{table}[h!]
\centering
\caption{Model state variables and parameters}
\label{variables}
\begin{tabular}{l l}
\hline \rowcolor{Gray}
\multicolumn{2}{l}{\textbf{State Variables}} \\
\hline
$x_1$ & Printer mount position (m) \\
$x_2$ & Frame position (m) \\
$x_3$ & Overall gantry position (m) \\
$x_4$ & Extruder carriage position (m) \\
$\theta_4$ & Pulley 4 Angular position (radians) \\
$\theta_5$ & Pulley 5 Angular position (radians) \\
\hline \rowcolor{Gray}
\multicolumn{2}{l}{\textbf{Model Parameters}} \\
\hline
$m_1$ & Printer mount mass (kg) \\
$m_2$ & Frame mass (kg) \\
$m_3$ & Overall gantry mass (kg) \\
$m_4$ & Pulley 4 mass (kg) \\
$m_5$ & Pulley 5 mass (kg) \\
$m_6$ & Extruder carriage mass (kg) \\
$k_1$ & Ground/mount stiffness (N/m) \\
$k_2$ & Frame/ground stiffness (N/m) \\
$k_3$ & Extuder carriage/frame stiffness (N/m) \\
$k_4$ & Section 1 belt stiffness (N/m) (length changes) \\
$k_5$ & Section 2 belt stiffness (N/m) (length changes) \\
$k_6$ & Section 3 belt stiffness (N/m) (length fixed) \\
$\beta_1$ & Ground/mount damping coefficient (N$\times$s/m) \\
$\beta_2$ & Frame/ground damping coefficient (N$\times$s/m) \\
$\beta_3$ & Extuder carriage/frame damping coefficient (N$\times$s/m) \\
$\beta_4$ & Section 1 belt damping coefficient (N$\times$s/m) (length changes) \\
$\beta_5$ & Section 2 belt damping coefficient (N$\times$s/m) (length changes) \\
$\beta_6$ & Section 3 belt damping coefficient (N$\times$s/m) (length fixed) \\
\hline
\end{tabular}
\end{table}

\section{DYNAMIC MODEL}
\subsection{System Model}
This work began by adapting and updating the dynamic FFF printer gantry model developed by Sharma and Patterson~\cite{Sharma2023A, Sharma2023B} and implementing a passive control system in it. In this model, the gantry can be thought of as a three-stage spring-mass-damper system with the three components being the printer mount (a desk or workbench), the frame, and the gantry system itself. This decomposition is shown in Figure~\ref{fig:system}. The power source is a motor mounted on the gantry itself and driving the extruder carriage using a belt. The main components of the gantry system itself are the extruder carriage, the rails, the two pulleys (one driven) and the three belt sections, two of which change in length throughout the operation. In this model, the build plate motion (Y-axis) is not considered and the Z-axis motion is assumed to be so slow that it does not introduce any significant vibration into the system. The Z position is important for calculating the model parameters for the frame, making this a 2-D model. The state variables are then all in the form of position with respect to time, as shown in Table~\ref{variables} along with the main system parameters.   

\subsection{Newton-Euler Decomposition}
As described by Sharma and Patterson~\cite{Sharma2023A, Sharma2023B}, the first step is to obtain the conservative dynamic model using a Newton-Euler decomposition. This model does not include any damping forces. Following the system model described in the last section, the gantry consists of three main components: The base of the machine (bench or work surface connected to the ground), the frame, and the gantry itself. This is demonstrated in Figure~\ref{fig:system} as a spring-mass-damper system. The gantry system will also have motion from the belts, pulleys, and extruder carriage. The force $D(t)$ is the time-variant force put into the whole system by the motion of the components within the gantry itself. The mass $m_3$ will be the mass of the entire extruder system, with the simplifying assumption that the belts mass is trivial. From this, conservative equations of motion can be found and are shown in Equations~\ref{eq1}-\ref{eq3}.       

\begin{footnotesize}
\begin{equation}
\label{eq1}
\ddot{x}_1 = -\frac{k_1 + k_2}{m_1}x_1 + \frac{k_2}{m_1}x_2
\end{equation}
\begin{equation}
\label{eq2}
\ddot{x}_2 = \frac{k_2}{m_2}x_1-\frac{k_2 + k_3}{m_2}x_2+\frac{k_3}{m_2}x_3
\end{equation}
\begin{equation}
\label{eq3}
\ddot{x}_3 = \frac{k_3}{m_T}x_2 - \frac{k_3}{m_3}x_3 + \frac{D(t)}{m_3}
\end{equation}
\end{footnotesize}

Adding in the motion of the extruder carriage, belts, and pulleys (Figure~\ref{fig:system}) requires an analysis of the local motion during operations which translates into $D(t)$ from the perspective of the frame and mount. Note that all the input force is in the form of a nonconservative torque which is applied to one of the pulleys. For the purposes of this model, the assumption was made that the belts used would be standard glass fiber GT2 drive belts. The pulleys are only able to rotate relative to the gantry, but the entire gantry is able to translate during the motion of the printer. These equations are shown in Equations~\ref{eq4}-\ref{eq9}.      

\begin{footnotesize}
\begin{equation}
\label{eq4}
\ddot{x}_1 = -\frac{k_1 + k_2}{m_1}x_1 + \frac{k_2}{m_1}x_2
\end{equation}
\begin{equation}
\label{eq5}
\ddot{x}_2 = \frac{k_2}{m_2}x_1-\frac{k_2 + k_3}{m_2}x_2+\frac{k_3}{m_2}x_3
\end{equation}
\begin{equation}
\label{eq6}
\ddot{x}_3 = \frac{k_3}{m_3}x_2 - \frac{k_3 + k_4 + k_5}{m_3}x_3 + \frac{k_4 + k_5}{m_3}x_4 + \frac{k_4R}{m_3}\theta_4 + \frac{k_5R}{m_3}\theta_5 
\end{equation}
\begin{equation}
\label{eq7}
\ddot{x}_4 = \frac{k_4+k_5}{m_6}x_3 - \frac{k_4 + k_5}{m_6}x_4 - \frac{k_4R}{m_6}\theta_4 - \frac{k_5R}{m_6}\theta_5 
\end{equation}
\begin{equation}
\label{eq8}
\ddot{\theta}_4 = \frac{2k_4}{m_4R}x_3 - \frac{2k_4}{m_4R}x_4 - \frac{2(k_4+k_6)}{m_4}\theta_4 + \frac{2k_6}{m_4}\theta_5
\end{equation}
\begin{equation}
\label{eq9}
\ddot{\theta}_5 = \frac{2k_5}{m_5R}x_3 - \frac{2k_5}{m_5R}x_4 + \frac{2k_6}{m_5}\theta_4 - \frac{2(k_5+k_6)}{m_5}\theta_6
\end{equation}
\end{footnotesize}

\subsection{Nonconservative Model}
To get the final dynamic model, it is necessary to add the nonconservative elements, which will be the friction and damping forces in the components. For this model, it is assumed that the printer is well maintained and that the bearings are in excellent condition, with the result that friction in the system is trivial and can be neglected. Therefore, the only non-conserative model elements are the damping forces. When considering damping forces, the equations of motion can be derived and are shown in Equations~\ref{eq13}-\ref{eq18}. Some other important model assumptions were related to the stiffnesses of the system components and the belt preload. Specifically, 

\begin{itemize}
\item $k_1$ and $k_2$ are constants and can be measured directly from the printer setup.
\item $k_3$ is a function of the height of the gantry and can be measured or calculated using a frame beam model. 
\item For the purposes of this model (and upon checking available printers) it was assumed that the belt preload $F_{pl}$ was 45N. 
\item $k_6$ is constant since its length does not change during operation. It is calculated from Wang et al.~\cite{Wang2018} as 
\begin{footnotesize}
\begin{equation}
\label{eq10}
k_6 = \frac{C_{sp}b}{L}+\frac{F_{pl}}{L_0} = \text{constant}
\end{equation}
\end{footnotesize}
where $C_{sp}$ is the characteristic stiffness of the belt, $b$ is the belt width, $L$ is the belt length between pulley contact points,and $L_0$ is the total belt length at the start of the printer motion. 
\item $k_4$ and $k_5$ were calculated from Wang et al.~\cite{Wang2018} as:
\begin{footnotesize}
\begin{equation}
\label{eq11}
k_4 = \frac{C_{sp}b}{L_1+x_4} + \frac{F_{pl}}{L_0}
\end{equation}
\begin{equation}
\label{eq12}
k_5 = \frac{C_{sp}b}{L_2-x_4} + \frac{F_{pl}}{L_0}
\end{equation}
\end{footnotesize}
where $L_1+x_4$ is the distance (as a function of time) from the extruder gantry to Pulley 4 and $L_2-x_4$ is the distance (as a function of time) from the extruder gantry to Pulley 5. For the preload component, $L_0$ was the total length of the belt when starting the printer motion. In these cases, $F_{pl}/L_0$ is a constant. 
\item The ideal motion of the system (no compliance or vibration) would be: 
\begin{equation}
T_{input}(t) = m_6 \ddot{x}_4 R + \frac{J_4 \ddot{x}_4}{R} + \frac{J_5 \ddot{x}_4}{R}
\end{equation}
which can be solved for $\ddot{x}_4$ and where $J_x$ is the moment of inertia for Pulley $x$, and $R$ is the pulley diamater, assuming all the pulleys are the same size.
\end{itemize} 

\begin{figure*}
\begin{footnotesize}
\begin{equation}
\label{eq13}
\ddot{x}_1 = -\frac{k_1 + k_2}{m_1}x_1 + \frac{k_2}{m_1}x_2 - \frac{\beta_1 + \beta_2}{m_1}\dot{x}_1 + \frac{\beta_2}{m_1}\dot{x}_2
\end{equation}
\begin{equation}
\label{eq14}
\ddot{x}_2 = \frac{k_2}{m_2}x_1-\frac{k_2 + k_3}{m_2}x_2+\frac{k_3}{m_2}x_3 + \frac{\beta_2}{m_2}\dot{x}_1 - \frac{\beta_2+\beta_3}{m_2}\dot{x}_2 + \frac{\beta_3}{m_2}\dot{x}_3
\end{equation}
\begin{equation}
\label{eq15}
\ddot{x}_3 = \frac{k_3}{m_3}x_2 - \frac{k_3 + k_4 + k_5}{m_3}x_3 + \frac{k_4 + k_5}{m_3}x_4 + \frac{k_4R}{m_3}\theta_4 + \frac{k_5R}{m_3}\theta_5 + \frac{\beta_3}{m_3}\dot{x}_2 - \frac{\beta_3 + \beta_4 + \beta_5}{m_3}\dot{x}_3 \\ + \frac{\beta_4+\beta_5}{m_3}\dot{x}_4 + \frac{\beta_4 R}{m_3}\dot{\theta}_4 + \frac{\beta_5 R}{m_3}\dot{\theta}_5 
\end{equation}
\begin{equation}
\label{eq16}
\ddot{x}_4 = \frac{k_4+k_5}{m_6}x_3 - \frac{k_4 + k_5}{m_6}x_4 - \frac{k_4R}{m_6}\theta_4 - \frac{k_5R}{m_6}\theta_5 + \frac{\beta_4 + \beta_5}{m_6}\dot{x}_3 - \frac{\beta_4 + \beta_5}{m_6}\dot{x}_4 - \frac{\beta_4 R}{m_6}\dot{\theta}_4 - \frac{\beta_5 R}{m_6}\dot{\theta}_5
\end{equation}
\begin{equation}
\label{eq17}
\ddot{\theta}_4 = \frac{2k_4}{m_4R}x_3 - \frac{2k_4}{m_4R}x_4 - \frac{2(k_4+k_6)}{m_4}\theta_4 + \frac{2k_6}{m_4}\theta_5 + \frac{2\beta_4}{m_4R}\dot{x}_3 - \frac{2\beta_4}{m_4R}\dot{x}_4 - \frac{2(\beta_4 + \beta_6)}{m_4}\dot{\theta}_4 + \frac{2\beta_6}{m_4}\dot{\theta}_5 + \frac{2T_{input}}{m_4R^2}
\end{equation}
\begin{equation}
\label{eq18}
\ddot{\theta}_5 = \frac{2k_5}{m_5R}x_3 - \frac{2k_5}{m_5R}x_4 + \frac{2k_6}{m_5}\theta_4 - \frac{2(k_5+k_6)}{m_5}\theta_6 + \frac{2\beta_5}{m_5R}\dot{x}_3 - \frac{2\beta_5}{m_5R}\dot{x}_4 + \frac{2\beta_6}{m_5}\dot{\theta}_4 - \frac{2(\beta_5 + \beta_6)}{m_5}\dot{\theta}_5
\end{equation}
\end{footnotesize}
\end{figure*}

\begin{figure*}
\begin{scriptsize}
\setcounter{MaxMatrixCols}{12}
\renewcommand*{\arraystretch}{1.5}
\setlength\arraycolsep{1.5pt}
\begin{equation}
\label{eq19}
[A] = 
\begin{bmatrix}
0 & 1 & 0 & 0 & 0 & 0 & 0 & 0 & 0 & 0 & 0 & 0 \\
-\frac{k_1+k_2}{m_1} & -\frac{\beta_1+\beta_2}{m_1} & \frac{k_2}{m_1} & \frac{\beta_2}{m_1} & 0 & 0 & 0 & 0 & 0 & 0 & 0 & 0 \\
0 & 0 & 0 & 1 & 0 & 0 & 0 & 0 & 0 & 0 & 0 & 0 \\
\frac{k_2}{m_2} & \frac{\beta_2}{m_2} & -\frac{k_2 + k_3}{m_2} & - \frac{\beta_2+\beta_3}{m_2} & \frac{k_3}{m_2} & \frac{\beta_3}{m_2} & 0 & 0 & 0 & 0 & 0 & 0  \\
0 & 0 & 0 & 0 & 0 & 1 & 0 & 0 & 0 & 0 & 0 & 0  \\
0 & 0 & \frac{k_3}{m_3} & \frac{\beta_3}{m_3} & -\frac{k_3 + k_4 + k_5}{m_3} & -\frac{\beta_3 + \beta_4 + \beta_5}{m_3} & \frac{k_4 + k_5}{m_3} & \frac{\beta_4+\beta_5}{m_3} & \frac{k_4R}{m_3} & \frac{\beta_4 R}{m_3} & \frac{k_5R}{m_3} & \frac{\beta_5 R}{m_3} \\
0 & 0 & 0 & 0 & 0 & 0 & 0 & 1 & 0 & 0 & 0 & 0 \\
0 & 0 & 0 & 0 & \frac{k_4+k_5}{m_6} & \frac{\beta_4 + \beta_5}{m_6} & - \frac{k_4 + k_5}{m_6} & - \frac{\beta_4 + \beta_5}{m_6} & -\frac{k_4R}{m_6} & -\frac{\beta_4 R}{m_6} & - \frac{k_5R}{m_6} & - \frac{\beta_5 R}{m_6} \\
0 & 0 & 0 & 0 & 0 & 0 & 0 & 0 & 0 & 1 & 0 & 0 \\
0 & 0 & 0 & 0 & \frac{2k_4}{m_4R} & \frac{2\beta_4}{m_4R} & - \frac{2k_4}{m_4R} & - \frac{2\beta_4}{m_4R} & - \frac{2(k_4+k_6)}{m_4} & - \frac{2(\beta_4 + \beta_6)}{m_4} & \frac{2k_6}{m_4} & \frac{2\beta_6}{m_4} \\
0 & 0 & 0 & 0 & 0 & 0 & 0 & 0 & 0 & 0 & 0 & 1 \\
0 & 0 & 0 & 0 & \frac{2k_5}{m_5R} & \frac{2\beta_5}{m_5R} & - \frac{2k_5}{m_5R} & - \frac{2\beta_5}{m_5R} & \frac{2k_6}{m_5} & \frac{2\beta_6}{m_5} & - \frac{2(k_5+k_6)}{m_5} & - \frac{2(\beta_5 + \beta_6)}{m_5} \\
\end{bmatrix}
\end{equation}
\setlength\arraycolsep{5pt}
\begin{equation}
\label{eq20}
[B] = 
\begin{bmatrix}
0 & 0 & 0 & 0 & 0 & 0 & 0 & 0 & 0 & \frac{2}{m_4R^2} & 0 & 0
\end{bmatrix}^T
\end{equation}
\setlength\arraycolsep{5pt}
\begin{equation}
\label{eq21}
[C] = 
\begin{bmatrix}
0 & 0 & 0 & 0 & 0 & 0 & 1 & 1 & 0 & 0 & 0 & 0
\end{bmatrix}^T
\end{equation}
\setlength\arraycolsep{3pt}
\begin{equation}
\label{eq22}
[X] = 
\begin{bmatrix}
z_1 & z_2 & z_3 & z_4 & z_5 & z_6 & z_7 & z_8 & z_9 & z_{10} & z_{11} & z_{12}
\end{bmatrix}^T
\end{equation}
\begin{equation}
\label{eq23}
u = T_{input}
\end{equation}
\end{scriptsize}
\end{figure*}

\subsection{State Space Model}
After completion of the motion equations, the state derivatives can be taken to derive the state space model: $\dot{z}_1 = \dot{x}_1 = z_2$, $\dot{z}_2 = \ddot{x}_1$, $\dot{z}_3 = \dot{x}_2 = z_4$, $\dot{z}_4 = \ddot{x}_2$, $\dot{z}_5 = \dot{x}_3 = z_6$, $\dot{z}_6 = \ddot{x}_3$, $\dot{z}_7 = \dot{x}_4 = z_8$, $\dot{z}_8 = \ddot{x}_4$, $\dot{z}_9 = \dot{\theta}_4 = z_{10}$, $\dot{z}_{10} = \ddot{\theta}_4$, $\dot{z}_{11} = \dot{\theta}_5 = z_{12}$, $\dot{z}_{12} = \ddot{\theta}_5$. The resulting state-space model is shown in Equations~\ref{eq13}-\ref{eq18}. The model represents $\dot{X}=AX + Bu$, $Y=CX+Du$ where $A$ is shown in Equation~\ref{eq19}, $B$ is shown in Equation~\ref{eq20}, $C$ is shown in Equation~\ref{eq21}, $X$ is shown in Equation~\ref{eq22}, and $u$ is shown in Equation~\ref{eq23}. Note that Equations~\ref{eq10}-\ref{eq12} were not integrated directly into the equation for this paper due to space constraints but were included in all models and simulations shown.

\begin{figure}[h!]
\centering
\includegraphics[width=0.45\textwidth]{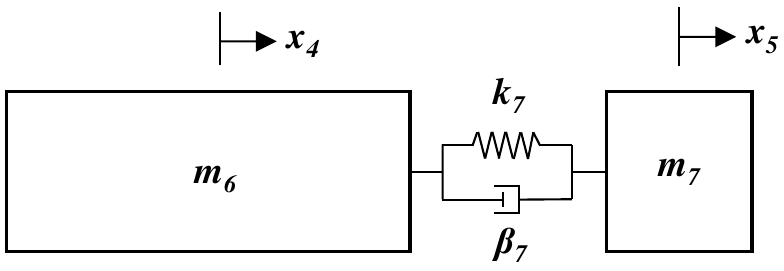}
\caption{\textit{Passive controller as a spring-mass-damper system which can be tuned to counteract the natural vibration of the system}}
\label{fig:SRD1}
\end{figure}

\section{PASSIVE CONTROL ON EXTRUDER}
A possible approach for a passive control system is to attach a tuned spring-mass-damper system to the extruder carriage, with the objective of counteracting the vibration of the regular system. This concept was previously presented by Sharma and Patterson~\cite{Sharma2023B, Sharma2023A}, but the current paper presents a fully developed, improved, and more realistic model. Since this is the addition of extra mass and vibration to the system, this approach may also make the problem worse if the tuning is not done correctly. For this initial study, the objective was to study the effects of adding the new system, with the specifics of tuning reserved for future work. The spring-mass-damper model is shown in Figure~\ref{fig:SRD1}. The updated equations of motion are the same as for the basic model with the exception of $\ddot{x}_4$ and the addition of an $\ddot{x}_5$ equation. These updated equations are as follows: 

\begin{scriptsize}
\begin{equation}
\label{eq24}
\begin{split}
\ddot{x}_4 = \frac{k_4+k_5+k_7}{m_6}x_3 - \frac{k_4 + k_5}{m_6}x_4 - \frac{k_4R}{m_6}\theta_4 - \frac{k_5R}{m_6}\theta_5 \\ + \frac{\beta_4 + \beta_5}{m_6}\dot{x}_3 - \frac{\beta_4 + \beta_5+\beta_7}{m_6}\dot{x}_4 - \frac{\beta_4 R}{m_6}\dot{\theta}_4 - \frac{\beta_5 R}{m_6}\dot{\theta}_5 \\ + \frac{k_7}{m_6}x_5 + \frac{\beta_7}{m_6}\dot{x}_5
\end{split}
\end{equation}
\begin{equation}
\label{eq25}
\ddot{x}_5 = \frac{k_7}{m_7}x_4 - \frac{k_7}{m_7}x_5 + \frac{\beta_7}{m_7}\dot{x}_4-\frac{\beta_7}{m_7}\dot{x}_5
\end{equation}
\end{scriptsize}

\begin{figure*}
\setcounter{MaxMatrixCols}{14}
\renewcommand*{\arraystretch}{1.5}
\setlength\arraycolsep{-0.25pt}
\begin{tiny}
\begin{equation}
\label{eq26}
[A] = 
\begin{bmatrix}
0 & 1 & 0 & 0 & 0 & 0 & 0 & 0 & 0 & 0 & 0 & 0 & 0 & 0 \\
-\frac{k_1+k_2}{m_1} & -\frac{\beta_1+\beta_2}{m_1} & \frac{k_2}{m_1} & \frac{\beta_2}{m_1} & 0 & 0 & 0 & 0 & 0 & 0 & 0 & 0 & 0 & 0 \\
0 & 0 & 0 & 1 & 0 & 0 & 0 & 0 & 0 & 0 & 0 & 0 & 0 & 0 \\
\frac{k_2}{m_2} & \frac{\beta_2}{m_2} & -\frac{k_2 + k_3}{m_2} & - \frac{\beta_2+\beta_3}{m_2} & \frac{k_3}{m_2} & \frac{\beta_3}{m_2} & 0 & 0 & 0 & 0 & 0 & 0 & 0 & 0  \\
0 & 0 & 0 & 0 & 0 & 1 & 0 & 0 & 0 & 0 & 0 & 0 & 0 & 0 \\
0 & 0 & \frac{k_3}{m_3} & \frac{\beta_3}{m_3} & -\frac{k_3 + k_4 + k_5}{m_3} & -\frac{\beta_3 + \beta_4 + \beta_5}{m_3} & \frac{k_4 + k_5}{m_3} & \frac{\beta_4+\beta_5}{m_3} & \frac{Rk_4}{m_3} & \frac{\beta_4 R}{m_3} & \frac{k_5R}{m_3} & \frac{\beta_5 R}{m_3} & 0 & 0 \\
0 & 0 & 0 & 0 & 0 & 0 & 0 & 1 & 0 & 0 & 0 & 0 & 0 & 0 \\
0 & 0 & 0 & 0 & \frac{k_4+k_5}{m_6} & \frac{\beta_4 + \beta_5}{m_6} & - \frac{k_4 + k_5 + k_7}{m_6} & - \frac{\beta_4 + \beta_5 + \beta_7}{m_6} & -\frac{Rk_4}{m_6} & -\frac{\beta_4 R}{m_6} & - \frac{k_5R}{m_6} & - \frac{\beta_5 R}{m_6} & \frac{k_7}{m_6} & \frac{\beta_7}{m_6} \\
0 & 0 & 0 & 0 & 0 & 0 & 0 & 0 & 0 & 1 & 0 & 0 & 0 & 0 \\
0 & 0 & 0 & 0 & \frac{2k_4}{m_4R} & \frac{2\beta_4}{m_4R} & - \frac{2k_4}{m_4R} & - \frac{2\beta_4}{m_4R} & - \frac{2(k_4+k_6)}{m_4} & - \frac{2(\beta_4 + \beta_6)}{m_4} & \frac{2k_6}{m_4} & \frac{2\beta_6}{m_4} & 0 & 0 \\
0 & 0 & 0 & 0 & 0 & 0 & 0 & 0 & 0 & 0 & 0 & 1 & 0 & 0\\
0 & 0 & 0 & 0 & \frac{2k_5}{m_5R} & \frac{2\beta_5}{m_5R} & - \frac{2k_5}{m_5R} & - \frac{2\beta_5}{m_5R} & \frac{2k_6}{m_5} & \frac{2\beta_6}{m_5} & - \frac{2(k_5+k_6)}{m_5} & - \frac{2(\beta_5 + \beta_6)}{m_5} & 0 & 0 \\
0 & 0 & 0 & 0 & 0 & 0 & 0 & 0 & 0 & 0 & 0 & 0 & 0 & 1 \\
0 & 0 & 0 & 0 & 0 & 0 & \frac{k_7}{m_7} & \frac{\beta_7}{m_7} & 0 & 0 & 0 & 0 & -\frac{k_7}{m_7} & -\frac{\beta_7}{m_7} \\
\end{bmatrix}
\end{equation}
\end{tiny}
\begin{scriptsize}
\setlength\arraycolsep{5pt}
\begin{equation}
\label{eq27}
[B] = 
\begin{bmatrix}
0 & 0 & 0 & 0 & 0 & 0 & 0 & 0 & 0 & \frac{2}{m_4R^2} & 0 & 0 & 0 & 0
\end{bmatrix}^T
\end{equation}
\setlength\arraycolsep{5pt}
\begin{equation}
\label{eq28}
[C] = 
\begin{bmatrix}
0 & 0 & 0 & 0 & 0 & 0 & 1 & 1 & 0 & 0 & 0 & 0 & 0 & 0
\end{bmatrix}^T
\end{equation}
\setlength\arraycolsep{3pt}
\begin{equation}
\label{eq29}
[X] = 
\begin{bmatrix}
z_1 & z_2 & z_3 & z_4 & z_5 & z_6 & z_7 & z_8 & z_9 & z_{10} & z_{11} & z_{12} & z_{13} & z_{14}
\end{bmatrix}^T
\end{equation}
\begin{equation}
\label{eq30}
u = T_{input}
\end{equation}
\end{scriptsize}
\end{figure*}

\noindent Adding this model to the system adds another state variable ($x_5$) and two more state derivatives ($\dot{z}_{13}=\dot{x}_5=z_{14}$ and $\dot{z}_{14} = \ddot{x}_5$). The resulting new state space model is shown in Equations~\ref{eq26}-\ref{eq30}.

\renewcommand*{\arraystretch}{1.05}
\begin{table}[b!]
\centering
\caption{Case study parameters using orthogonal design of experiments.}
\label{DOE}
\begin{tabular}{l l l l}
\hline \rowcolor{Gray}
\textbf{Case number} & \textbf{$m_7$ ($kg$)} & \textbf{$k_7$ ($N/m$)} & \textbf{$\beta_7$ ($Ns/m$)} \\
\hline
Passive case 1 & 0.005 & 1 & 0.1 \\
Passive case 2 & 0.005 & 50 & 0.5 \\
Passive case 3 & 0.005 & 100 & 1 \\
Passive case 4 & 0.05 & 1 & 0.5 \\
Passive case 5 & 0.05 & 50 & 1 \\
Passive case 6 & 0.05 & 100 & 0.1 \\
Passive case 7 & 0.5 & 1 & 1 \\
Passive case 8 & 0.5 & 50 & 0.1 \\
Passive case 9 & 0.5 & 100 & 0.5 \\
\hline
\end{tabular}
\end{table}

\begin{figure*}[h!]
\centering
\includegraphics[width=1\textwidth]{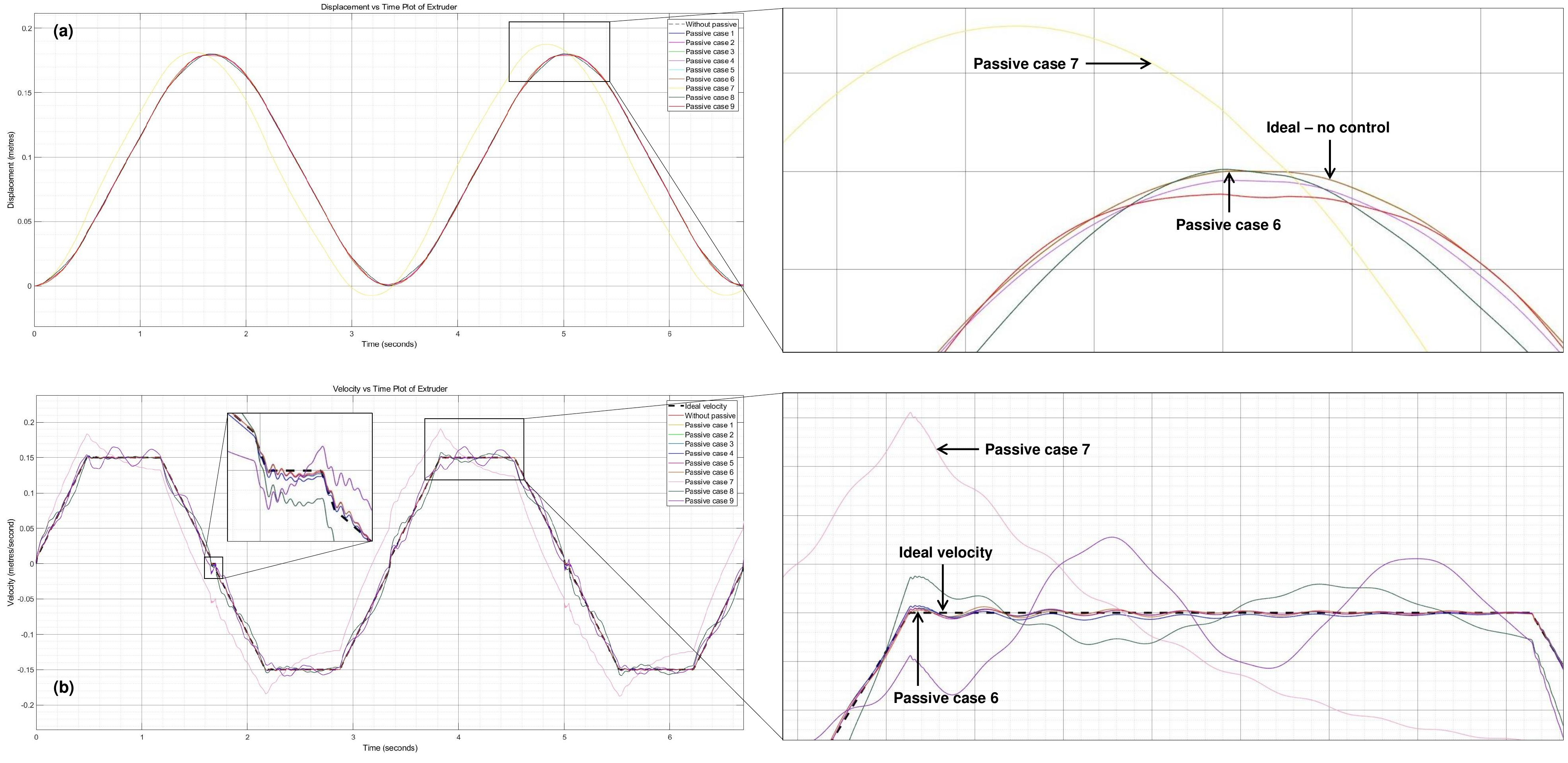}
\caption{\textit{Position and velocity plots with the passive spring-mass-damper system attached for the none text cases.}}
\label{fig:Plot}
\end{figure*}

\section{CASE STUDY}
To explore the impact of adding the passive spring-mass-damper system to the extruder carriage, as case study was conducted in which the three parameters (mass, spring stiffness, damping coefficient) were varied using an orthogonal design of the experiments. The parameters of the system are shown in Table~\ref{DOE}. The motion of the printer was modeled compared with the ideal motion. The $T_{input}$ was modeled as a ramp-up with realistic machine parameters: The maximum acceleration was 1000 $mm/s^2$, the print acceleration was 300 $mm/s^2$, the print velocity was 150 $mm/s$, the jerk was 10 $mm/s^2$, the print distance (each way) was 180 $mm$, and the Z hop time was 0.04 $s$. The basic parameters used were $m_1 = 500~kg$, $m_2 = 6.201~kg$, $m_3 = 0.721~kg$, $m_4 = 0.017~kg$ (non-driven), $m_5 = 0.019~kg$ (driven), $m_6 = 0.611~kg$, $R = 8~mm$, $L_1 = 80~mm$, $b = 6~mm$, $\beta_1 = 10,000~Ns/m$, $\beta_2 = 1,000~Ns/m$, $\beta_3 = 1~Ns/m$, $k_1 = 10^6~N/m$, $k_2 = 10^5~N/m$, $k_3 = 6410~N/m$, $C_{sp} = 1.74 \times 10^6~N/m$, $L = 350~mm$, $L_0 = 750~mm$, $\beta_4 = 5~Ns/m$, $\beta_5 = 5~Ns/m$, $\beta_6 = 5~Ns/m$. All of these parameters were generated by measuring a real FFF machine available in the lab. The calculations were performed using Matlab/Simulink. The position and velocity curves for each set of passive controller parameters are shown in Figure~\ref{fig:Plot}a and Figure~\ref{fig:Plot}b, respectively. As shown in the plots, the addition of this passive system has a large effect, driven mainly by the combinations of parameters. For both position ($x_4$) and velocity ($\dot{x}_4$), passive case 6 ($m_7 = 0.5~kg$, $k_7 = 100~N/m$, and $\beta_7 = 0.5~Ns/m$) produced the best results and provided a position curve almost the same as the ideal, while passive case 7 ($m_7 = 0.5~kg$, $k_7 = 1~N/m$ and $\beta_7 = 1~Ns/m$) produced the worst results. For velocity, these conclusions were valid in both of the velocity transition regions shown in Figure~\ref{fig:Plot}b.         

\section{HYBRID ACTIVE-PASSIVE CONTROL}
\begin{figure}[h!]
\centering
\includegraphics[width=0.5\textwidth]{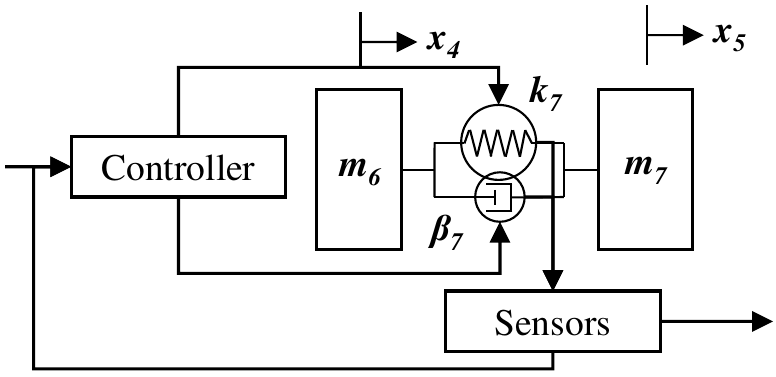}
\caption{\textit{Hybrid active-passive control concept for the extruder carriage.}}
\label{fig:F4}
\end{figure}

From the case study, it was concluded that proper parameter tuning is vital, as the parameters can improve the vibration issues for position and velocity, but can also make them much worse. The parameter tuning can be done carefully with the main printer dynamics in mind for a full-passive system. However, it was also noted that this kind of control setup could be used with a hybrid active-passive system. Instead of directly controlling the motion of the printer gantry (which would require actuators and severely restrict the gantry motion), the value of $k_7$ and $\beta_7$ could be controlled using actuators within the additional spring-mass-damper system and be controlled much more easily with a self-contained system. This concept is demonstrated in Figure~\ref{fig:F4}. Other previous work has shown the value of using active stiffening and damping as a control system~\cite{Allison2014, Herber2018, Cui2019}, so it is clearly a future useful option for FFF machines, allowing them to print faster and more precisely. A system for collecting real-time information for a feedback loop is still needed, but various vibration and position sensors could be used with this system which would be infeasible for a regular gantry as they would interfere with its movements.  

\section{CONCLUSIONS}
In this work, the FFF machine ganrty model developed by Sharma and Patterson~\cite{Sharma2023A, Sharma2023B} was updated and adapted for use with a passive vibration control system to allow faster and more precise printing. A case study showed the feasibility of this approach and showed some of the effects of changing the parameters on this passive system. This approach was shown to be used for future active-passive hybrid systems, which would be used to counteract vibration in the system without interfering with the mechanical motion of the printer gantry. The results of this work are useful for the further development of FFF machines and other similar positioner-based systems such as cranes, pick-and-place machines, and various robotic systems, both for the design of the systems themselves and the manufacturing of precice componets needed by the systems.

\section*{ACKNOWLEDGMENTS}
The authors thank Dr. James T. Allison for his ideas and help in developing an earlier version of this research approach. No external funding was used to produce this work. All conclusions and claims are solely those of the named authors. All code, models, and raw data for this article are available upon reasonable request from the corresponding author. 


\end{document}